\documentclass[10pt, conference]{ieeeconf}
\IEEEoverridecommandlockouts
\usepackage{booktabs}
\usepackage{graphicx} 
\usepackage{amsmath}
\usepackage{amssymb}
\usepackage{xcolor}
\usepackage{hyperref}
\usepackage{pifont}

\newcommand{\ms}{\mbox{ManiSkill3}}
\newcommand{\rpd}{RPD}

\newcommand{\rfm}{VLA}
\newcommand{\vla}{VLA}

\title{\LARGE \bf
Refined Policy Distillation: From \vla{} Generalists to RL Experts
}

\author{Tobias Jülg$^{1}$, Wolfram Burgard$^{1}$ and Florian Walter$^{1}$
\thanks{$^{1}$All authors are with the Department of Computer Science~\& Artificial Intelligence, University of Technology Nuremberg, Germany. Contact: \texttt{[tobias.juelg, wolfram.burgard, florian.walter]@utn.de}}%
}

\begin{document}

\maketitle
\thispagestyle{empty}
\pagestyle{empty}

\begin{abstract}
  Vision-Language-Action Models (\rfm{}s) have demonstrated remarkable generalization capabilities in real-world experiments. However, their success rates are often not on par with expert policies, and they require fine-tuning when the setup changes.
  In this work, we introduce \emph{Refined Policy Distillation} (RPD), a novel Reinforcement Learning (RL)-based \emph{policy refinement} method that bridges this performance gap through a combination of on-policy RL with behavioral cloning. The core idea of RPD is to distill and refine \rfm{}s into compact, high-performing expert policies by guiding the student policy during RL exploration using the actions of a teacher \rfm{}, resulting in increased sample efficiency and faster convergence. We complement our method by fine-tuned versions of Octo and OpenVLA for \ms{} to evaluate RPD in simulation. While this is a key requirement for applying RL, it also yields new insights beyond existing studies on VLA performance in real-world settings. Our experimental results across various manipulation tasks show that RPD enables the RL student to learn expert policies that outperform the VLA teacher in both dense and sparse reward settings, while also achieving faster convergence than the RL baseline. Our approach is even robust to changes in camera perspective and can generalize to task variations that the underlying \rfm{} cannot solve. Our code, dataset, VLA checkpoints, and videos are available at \href{https://refined-policy-distillation.github.io/}{https://refined-policy-distillation.github.io}
\end{abstract}

\section{Introduction}
\bstctlcite{IEEEexample:BSTcontrol}

Recently, robot learning has seen a paradigm shift from Reinforcement Learning (RL) of task-specific policies to large generalist Vision-Language-Action Models (\rfm{}s) that are trained in a supervised fashion using imitation learning techniques~\cite{gato, rt1, rt2, octo, pizero, palme, openvla, gr2}.
\rfm{}s are easy to train and show good performance on different real-world manipulation tasks with a single model.
However, they usually need to be trained on a large number of human expert demonstrations, limiting their performance to the amount and variety of the training data.
Different from the field of natural language processing, creating more data is very cost-intensive in robotics, as it usually means collecting trajectories by humans through time-consuming teleoperation.
Due to limited training data, VLAs have problems generalizing across setups~\cite{vla_gen_gap}.
Applying \rfm{}s in new environments, thus, often involves fine-tuning on domain-specific data, requiring human effort yet again to record demonstrations, which further limits their widespread adoption.
Only recently, Xu \emph{et al.}~\cite{rldg} utilized data generated by RL agents for VLA fine-tuning, which can significantly speed up the adaptation to specific setups.

RL can, in principle, scale much better than imitation learning as an agent learns to maximize rewards through autonomous interaction with its environment and, thus, generates its own data.
Often, training is performed in simulation, which can be massively parallelized~\cite{mujoco, nvidia_gpu_sim, sapien}.
RL policies are also considerably more compact than VLAs as they have fewer parameters, resulting in faster inference and increased efficiency.

However, applying RL in robotics also poses several challenges:
Training is very sensitive to the hyperparameter configuration and often needs environment-specific tuning~\cite{hard_rl}.
RL agents require extensive exploration of the state space, which is usually sample-inefficient and precludes execution on a real robot.
Simulated environments are, therefore, essential for most RL algorithms to work.
The resulting \mbox{sim-to-real} gap can be closed by subsequent fine-tuning on the real robot.

\begin{figure}[t]
  \centering
  \includegraphics[width=\columnwidth]{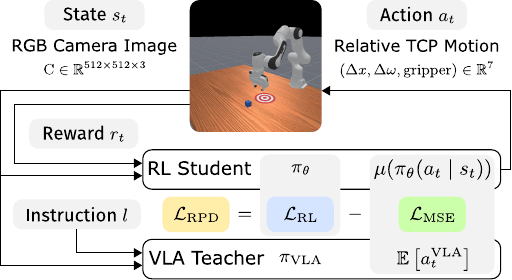}
  \caption{The architecture of \rpd{}:
We distill a \vla{} teacher into an RL student policy. This has two effects: First, the RL student agent is bootstrapped with guided exploration through the \vla{} teacher. Second, the RL student policy can interact with the environment and is, thus, able to surpass the \vla{}'s performance, refining the distilled policy.
  }
  \label{fig:arch}
\end{figure}

In this work, we introduce \emph{Refined Policy Distillation}~(RPD), a novel method for distilling and optimizing expert policies from  \vla{}s by combining on-policy RL with behavioral cloning.
The idea behind \rpd{} is that a \vla{} trained on large-scale datasets stores general task knowledge that can inform the training process of an RL algorithm, even if the policy by itself can solve the task only at a low or possibly even zero success rate.
Our evaluation shows that \rpd{} is not only capable of distilling and improving expert policies from fine-tuned \vla{}s but that it can also successfully distill \vla{} knowledge while learning unseen tasks.
This also applies when the camera angle changes, a situation in which current \vla{}s largely fail.
\rpd{} extends the Proximal Policy Optimization (PPO) algorithm~\cite{ppo} and only requires a small modification of its loss function.

\textbf{In summary, we make the following contributions:}
\begin{enumerate}
    \item We introduce RPD, a novel method for distilling and refining compact task-specific expert policies from large VLAs with on-policy RL.
    \item We demonstrate RPD's increased sample efficiency and stability on six different \vla{} in-distribution tasks for both dense and sparse reward settings.
    \item We demonstrate generalization to two out-of-distribution task variations and a changed camera perspective.
    \item We provide insight into how current \vla{}s perform in simulation by evaluating RPD with Octo~\cite{octo} and OpenVLA~\cite{openvla} variants fine-tuned for ManiSkill3.
\end{enumerate}

\section{Related Work}

In recent years, several approaches for learning generalist robot policies conditioned on vision and language inputs have been developed.
One of the most common ways is to use an existing VLM backbone trained on internet-scale data and fine-tune it on recorded robot trajectories~\cite{rt2, pizero, openvla, gr2}, most notably from the \mbox{Open X}-Embodiment dataset~\cite{rtx}.
Other models are based on large transformers and trained on vision and language or goal-image instructions from scratch~\cite{gato, rt1, octo, crossformer, vima}.
We refer to both of these transformer-based policy types as VLAs.

Not all \vla{}s are publicly available~\cite{rt2}, and for some, only the weights have been released, but not the source code~\cite{rt1}.
This precludes fine-tuning them on new tasks or setups.
However, both the code and the weights of many recent models are openly available~\cite{octo, pizero, openvla, crossformer}.
In this work, we use Octo~\cite{octo} and OpenVLA~\cite{openvla} for our tests as they represent two important classes of models, but \rpd{} works with any other \vla{}.

Policy Distillation (PD) is a widely investigated method in RL to distill teacher policies into student policies to increase training speed and decrease the complexity of the resulting policies~\cite {dpd}.
Early work in this domain investigated the distillation of DQN agents~\cite{pd}. The authors compared different approaches, including KL divergence and MSE loss between the Q functions.
KL divergence was used to distill DQN policies into PPO policies, too~\cite{green2019}.
PD can also be framed as a supervised policy transfer problem that allows the distillation from a teacher that uses a different RL algorithm than the student~\cite{reincarnatingrl}.
Compared to our method, those works only distill from RL teacher policies but not from generalist policies.

A recent work in this domain is Proximal Policy Distillation (PPD)~\cite{ppd}, which distills a PPO teacher policy~\cite{ppo} into a potentially larger PPO student policy.
PPD only samples from the student distribution, which allows it to lower the exploration bias induced by the teacher.
To incorporate the teacher's policy, PPD modifies the PPO loss by adding the KL divergence between the teacher's and the student's policy distribution together with a clipping term.
This method is transferable to \vla{} teacher policies as long as they are stochastic.
However, sampling the action distributions from a \vla{}, which is required for estimating the KL divergence, is compute-intensive and multiplies GPU load and memory usage by the sample size.
In this work, we employ a simple MSE loss, which avoids compute-intensive sampling of actions from a large-scale VLA.

Behavioral Cloning (BC) is an umbrella term for offline imitation learning: Given a dataset of state transitions, the goal is to learn a policy that follows the trajectories from the dataset as closely as possible~\cite{bc_pomerleau}.
Exact formulations differ, but the main idea is to maximize the likelihood of the state transitions in the dataset.
BC was also combined with RL to speed up the agent's exploration through a fixed pre-recorded dataset~\cite{dextman, ppo_bc}.
Recently, the term Offline RL has been used to describe similar approaches but they focus mainly on off-policy RL methods~\cite{offlinerl, rldp}.
Offline RL is related to PD as both approaches aim to learn from expert knowledge but uses datasets instead of learned policies.
We employ methods similar to BC to process external actions in the RL training loop but leverage pre-trained generalist policies to generate them.

Xu \emph{et al.}~\cite{rldg} investigated dataset generation with an offline soft actor-critic variant~\cite{rldp, sac} for VLA fine-tuning.
While their experiments show performance improvements of the \vla{} policy of up to 40\%, the method still requires already trained RL expert policies.
By contrast, our method distills large foundation models into smaller on-policy RL agents, which avoids training RL policies from scratch.
Through interaction with the environment, these policies can surpass the performance of their \vla{} teachers on their specific tasks and become expert policies.

In summary, compared to these previous works, our approach distills from large \vla{} models using RL. The resulting policies are compact and fast expert policies.
In all of our experiments, the resulting performance exceeds the \vla{}'s performance, which we refer to as \emph{policy refinement}.

\section{Methodology}\label{sec:meth}
\rpd{} is a method for distilling tasks from large generalist VLA policies into small and fast task-specific expert policies.
Standard supervised BC techniques usually learn policies that do not exceed the performance of the teacher. For this reason, we combine policy distillation with RL.
This helps the student agent to explore more efficiently and reach higher rewards faster, even in sparse reward settings. In addition, it also allows for refinement of the teacher policy to surpass its performance and become an expert policy for the distilled task. At the core of \rpd{} is a PPO RL agent with a modified objective function that includes a Mean Squared Error~(MSE) term between the action mean predicted by PPO and the expectation of the \vla{} action to pull the action mean of the RL policy closer to the predictions of the \vla{}.
\autoref{fig:arch} provides an overview of how \rpd{} integrates VLA actions into the agent-environment RL training loop.

\subsection{Refined Policy Distillation}
\rpd{} combines RL with a BC objective.
For RL, we model the interaction between an agent and its environment as a finite Markov Decision Process~(MDP)\linebreak $\mathcal{M} = (\mathcal{S}, \mathcal{A}, r, p, \rho_0, \gamma)$~\cite{sutton}. $\mathcal{S}$ is the state space, $\mathcal{A}$ the action space, $r: \mathcal{S} \times \mathcal{A}\rightarrow \mathbb{R}$ the reward function,\linebreak $p(s_{t+1} \mid s_t, a_t)$ the state transition probability distribution of the environment, $\rho_0$ the probability distribution over the initial states $s_0$, and $\gamma\in[0, 1]$ the discount factor.
Every task has its own MDP associated with it.
We consider stochastic policies $\pi_\theta(a_t \mid s_t)$ that are parameterized by weights $\theta$.
The goal of RL is to find a set of weights $\theta^*$ that maximizes the expected sum of discounted rewards:
\begin{align}
  \theta^* = \arg\max_\theta \mathbb{E}\left[\sum_{t=0}^{T-1}\gamma^t r(s_t, a_t)
  \right]
\end{align}
with $a_t \sim \pi_\theta(a_t \mid s_t)$, $s_0 \sim \rho_0$, $s_{t+1} \sim p(s_{t+1} \mid s_t, a_t)$, and the episode length $T$.

BC trains a policy from expert demonstrations that are usually provided as a dataset $\rho_\mathcal{D}$ of state transitions. We adopt a BC objective~\cite{dextman} that maximizes the likelihood of the policy under $\rho_\mathcal{D}$:
\begin{align}
  \max_\theta \sum_{(s, a)\in\rho_\mathcal{D}} \ln \pi_\theta (a\mid s).
  \label{eq:bc}
\end{align}
The transitions can also be sampled online and do not need to be stored in a dataset.
In our case they come from the \vla{} policy $\pi_{\text{\vla{}}}$.
Under the assumption of Gaussian action distributions, the BC objective essentially boils down to an MSE minimization objective between the two distributions' means and a term for the variance that we assume to be constant and therefore neglect.

The \rpd{} objective combines PPO with BC through a modified objective function:
\begin{align}
\mathcal{L}_{\text{RPD}}(\theta) = \mathcal{L}_\text{RL}(\theta) - \mathcal{L}_\text{MSE}(\theta) \label{eq:loss}
\end{align}
where $\mathcal{L}_{\text{RL}}$ is the standard PPO clipped surrogate objective function
\begin{align}
    \mathcal{L}_{\text{RL}}(\theta) = \mathbb{E}_t \left[ \min(r_t(\theta) \hat{A}_t, c( r_t(\theta), 1 - \epsilon, 1 + \epsilon) \hat{A}_t ) \right]
\end{align}
with clip function $c$ and the notation from Schulman \emph{et al.}~\cite{ppo}. $\mathcal{L}_{\text{MSE}}$ is the expectation of the MSE between the action mean $\mu(\pi_\theta(a_t \mid s_t))$ predicted by the PPO policy and the expectation $\mathbb{E}\left[a^\text{VLA}_t\right]$ of the actions $a^\text{VLA}_t \sim \pi_\text{VLA}(a_t \mid s_t)$ returned by the VLA policy:
\begin{align}
\mathcal{L}_\text{MSE}(\theta) = \mathbb{E}_t \left[ \left(\mu(\pi_\theta(a_t \mid s_t)) - \mathbb{E}\left[a^\text{VLA}_t\right]\right)^2 \right]
\end{align}
Note that the PPO objective is maximized while the MSE is minimized, which is why it appears with a negative sign.
In practice, we found that sampling a single VLA action per step to estimate the expectation already yields good performance. Increasing the sample size would slow down the training considerably and, therefore, was not further investigated.
Our training objective is to find a set of optimal weights $\theta^*$ that maximize the objective function $\mathcal{L}_{\text{RPD}}$:
\begin{align}
\theta^* = \arg\max_{\theta}\mathcal{L}_{\text{RPD}}(\theta)
\end{align}
In addition to the MSE-based RPD objective, which we refer to as RPD-MSE, we will also consider variants where the MSE loss is replaced by other loss functions.

\subsection{Tasks, Dataset, and VLA Fine-Tuning}
\begin{table*}
\centering
\vspace{5.2pt}
\caption{
Evaluation rewards (RW) and success rates (SR) of the VLAs on the selected \ms{} tasks.
}
\label{tab:tasks}
\resizebox{\linewidth}{!}{%
\begin{tabular}{l||l|c|rr|rr|rr|rr}
\toprule
Task ID &  Language Instruction & Fine-tuned & \multicolumn{2}{c}{openvla-base} & \multicolumn{2}{c}{openvla-finetuned} & \multicolumn{2}{c}{octo-base} & \multicolumn{2}{c}{octo-finetuned} \\
& & & RW & SR & RW & SR & RW & SR & RW & SR \\
\midrule
\texttt{LiftPegUpright-v1} & ``lift the peg upright'' & \checkmark                  & 0.20 & 0\% & 0.46 & \textbf{12\%} & 0.20 & 0\% & 0.41 & 0\% \\
\texttt{PickCube-v1}       & ``pick up the cube'' & \checkmark                      & 0.06 & 0\% & 0.18 & 6\% & 0.06 & 0\% & 0.17 & \textbf{9\%} \\
\texttt{PullCube-v1}       & ``pull the cube towards the robot base'' &\checkmark   & 0.07 & 0\% & 0.27 & \textbf{92\%} & 0.07 & 0\% & 0.24 & 90\% \\
\texttt{PullCubeTool-v1}   & ``pull the cube by using the red tool'' &\ding{55}     & 0.10 & 0\% & 0.04 & \textbf{12\%} & 0.10 & 0\% & 0.03 & 10\% \\
\texttt{PushCube-v1}       & ``push the cube away from the robot base'' &\checkmark & 0.10 & 0\% & 0.12 & 27\% & 0.10 & 0\% & 0.16 & \textbf{67\%} \\
\texttt{RollBall-v1}       & ``push the ball'' &\checkmark                          & 0.02 & 0\% & 0.03 & 4\% & 0.02 & 0\% & 0.06 & \textbf{10\%} \\
\texttt{StackCube-v1}      & ``stack the red cube on the green cube'' &\checkmark   & 0.05 & 0\% & 0.13 & \textbf{3\%} & 0.05 & 0\% & 0.13 & 1\% \\
\texttt{PokeCube-v1}       & ``push the cube by using the blue tool'' &\ding{55}    & 0.06 & 0\% & 0.09 & \textbf{12\%} & 0.06 & 0\% & 0.08 & \textbf{12\%} \\
\bottomrule
\end{tabular}
}
\end{table*}

\begin{figure}[t]
  \centering
  \vspace{5.2pt}
  \includegraphics[width=\linewidth]{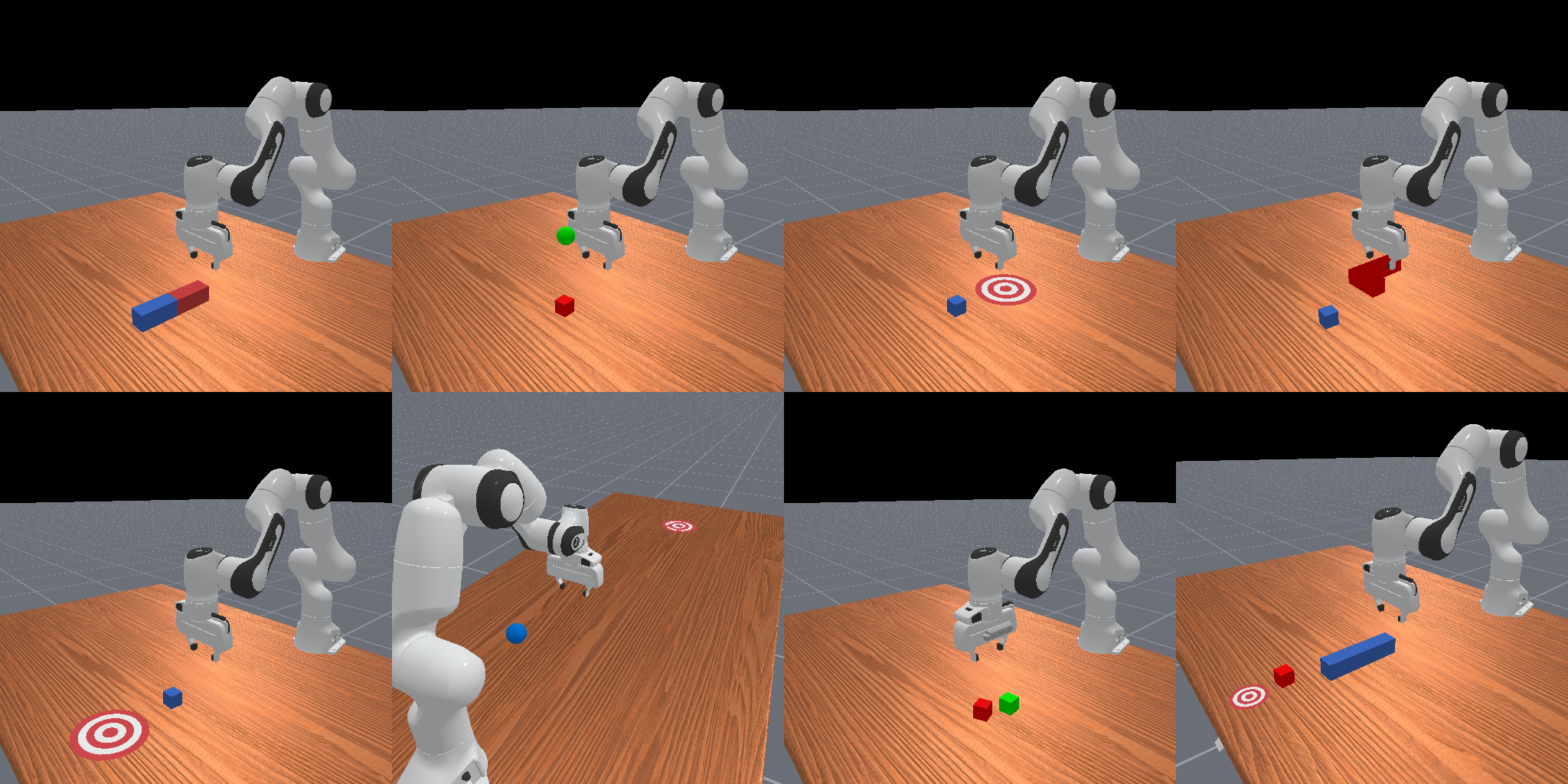}
  \caption{
    Overview of the eight different tasks from \ms{} that were used to distill expert policies with \rpd{}. See \autoref{tab:tasks} for the task names (tasks are depicted in row-major order).
  }
  \label{fig:envs}
\end{figure}

We use eight different manipulation tasks from \ms{}~\cite{maniskill} to evaluate \rpd{}.
\autoref{fig:envs} provides a visual overview.
Our criteria for choosing these tasks were the availability of demonstrations that are no longer than the maximum episode length and that all action spaces contain relative task space actions $a$:
\begin{equation}
    a = (\Delta x, \Delta \omega, \text{gripper})\in\mathbb{R}^7
\end{equation}
Since the effectiveness of guidance depends on the \vla{} teacher's performance, the student may not always benefit during training. In order to be able to appropriately assess the speedup enabled by RPD, we therefore focus our evaluation on tasks that are solvable by vanilla PPO.

The camera perspectives shown in \autoref{fig:envs} match the environment state observed by both the VLA and the RL agent.
Moreover, the same camera was used to record the VLA fine-tuning data.
\ms{} calls this the human camera.
It is normally not meant to be used for training as it contains visual hints such as a green sphere for the goal location of the cube in the PickCube task.

Goal locations in \ms{} are drawn randomly and provided to the agent only through explicit state information.
The VLAs considered in this work, however, only receive RGB camera state information and a fairly simple natural language instruction.
Therefore, at least one of these two input modalities needs to provide a hint for the currently selected target location.
While encoding the target in the language instruction appears to be a natural choice, the precision may be limited as VLAs are usually pre-trained on simple language instructions that do not contain exact spatial information.
We leverage the visual cues in \ms{}'s human camera to encode all relevant task information.
Another reason for using this camera is that its perspective shows the scene from the side rather than from the front. This side view corresponds better to the camera perspective used in many setups of the Open X-Embodiment dataset.
\\ \indent
Initially, we evaluated the published checkpoints for \mbox{OpenVLA} and Octo on the selected \ms{} tasks using the human camera and the language instructions shown in \autoref{tab:tasks}.
However, as reported in the table, both models had a success rate of zero on all tested tasks.
This real-to-sim gap is expected, as both Octo and OpenVLA were only trained on real-world datasets.
When using \ms{}'s raytracing renderer, the success rate of Octo increased to more than $2\%$ on PushCube, indicating that some positive domain transfer is possible for more realistic-looking input data.
Still, those success rates are too small for a teacher policy, and \ms{} does not support GPU rendering of the human camera with raytracing, which is required for RL.

In order to evaluate \rpd{} under realistic conditions, we first fine-tuned Octo and OpenVLA on the simulated environment using RL-generated expert demonstrations provided by \ms{}.
It is important to note that this step is only required because we evaluate \rpd{} in an environment these models were not trained on.
All demonstrations were rendered with the human camera at an image resolution of $256 \times 256$.
During training, we found that increasing the maximum episode length to 300 steps led to considerably higher success rates as the \ms{} dataset contains a subset with episodes longer than the maximum of the 50 steps used for most environments.
As long episodes can make RL algorithms behave less stable and also require a higher number of VLA inference steps, we excluded them from fine-tuning, which led to the expected success rates for both Octo and OpenVLA.
The results and language instructions for all tasks and models are summarized in \autoref{tab:tasks}.

The two tasks PullCubeTool and PokeCube are excluded from the fine-tuning dataset to test cross-task generalization.
While the normalized rewards are low, as expected, the reported success rates seem unreasonably high.
This is because in these two tasks, the agent is supposed to grasp a tool and use it to either pull or poke a cube.
We found that the VLAs were just pulling or pushing the cube to the target location without utilizing the tool.
This is counted as a success by the sparse reward implemented in \ms{} which just compares the distance towards the goal location and not whether the tool was actually used.
The dense reward, however, takes tool use into account and accurately shows the decreased performance.
Nevertheless, we still decided to keep both tasks.
Even if the learned behavior is just pushing and pulling, which the success rate can capture, it still differs from PushCube and PullCube because of distracting objects in the scene.
The fact that the language instruction is partly ignored by the VLA in that case is not relevant for testing \rpd{}'s generalization performance.

\subsection{VLA Integration Details}
To integrate different VLAs into the RL training loop, we defined a generic policy interface that takes a batch of visual observations and language instructions as input and returns a relative task space motion command including the gripper state.
To manage the dependencies of the different VLAs, we created separate Python environments and designed a generic VLA interface that is exposed via an HTTP API server.
During training, the RL agent connects to this server and sends batches of observations.
A new problem that comes with this approach is serialization.
Communication can quickly become a bottleneck when large batches of raw image data have to be serialized.
If we run both VLA inference and the RL training on the same machine, we use shared memory to avoid serialization.

Lastly, OpenVLA does not support batch processing in its inference code.
This limitation further slows down the already large model, as it complicates the implementation of parallel inference.
Moreover, the CPU-based preprocessing consumes substantial computional resources, requiring manual code optimization to offload it to the GPU.
Even after utilizing a parallelization factor of eight (running two \mbox{OpenVLA} instances on each GPU within a four Nvidia H100 node setup), it remains considerably slower than Octo operating on a single Nvidia A40 GPU.
Due to these constraints, we were only able to conduct a subset of our experiments using OpenVLA.

\section{Experimental Results}

We evaluated RPD on two VLAs, namely Octo~\cite{octo} and OpenVLA~\cite{openvla}.
The latter model, with its 7B parameters, is much larger than the former, with only 93M parameters, and therefore requires considerably more computing time, which is exacerbated by the lack of support for batched inference.
For this reason, we report evaluation results for OpenVLA on only a subset of the tasks.
Whenever the base model is not explicitly mentioned, we refer to Octo.

We split the evaluation into four parts:
First, we test RPD on the challenging PickCube task to explore the performance differences between loss variants and baselines.
We then evaluate RPD and its PPO baseline on a total of six different tasks that are included in the VLA fine-tuning dataset.
RL often struggles with sparse rewards due to inefficient exploration.
We argue that RPD guides this exploration process and, thus, should also improve sample efficiency and performance on sparse rewards.
To test this hypothesis, we also run the tasks mentioned above in a sparse reward setting.
Finally, we evaluate RPD's cross-task and cross-setup generalization capabilities by testing it on two tasks that it was not trained on and a variant of the PushCube task with an altered camera perspective.
The latter simulates small setup changes that can easily happen in reality and have a considerable effect on the performance of current VLMs.

\subsection{RPD Variants and Baselines}
We compare three different variants of RPD on \ms{}'s PickCube task:
\begin{enumerate}
    \item RPD-MSE, the main RPD variant as defined in \eqref{eq:loss}.
    \item RPD-L1, which uses an L1 loss instead of the MSE.
    \item RPD-BC, which uses a naive maximum likelihood loss as defined in \eqref{eq:bc}.
\end{enumerate}
We use the same policy architecture for all tasks, RPD variants, and baselines.
All RPD variants are trained with PPO and constant hyperparameters that were adopted without modification from the \ms{} baseline~\cite{maniskill}.
The only exception is the batch size, which had to be adjusted to fit the VLA on the same GPU as the RL training loop.
Due to computational constraints, we distilled from OpenVLA only with RPD-MSE.

We train policies with vanilla PPO and PPD~\cite{ppd} as baselines and use the PPO implementation from CleanRL~\cite{cleanrl} that is included in \ms{}. As there is no source code publicly available for PPD, we use our own implementation, which extends CleanRL's PPO. To compute the KL divergence required by PPD for its distillation loss, we sample ten actions from the VLA and fit a multivariate Gaussian distribution to it. All hyperparameters are set to \ms's defaults. The weighting factor of PPD is set to $\lambda=1.0$.

\begin{figure}[t]
  \centering
  \vspace{5.2pt}
  \includegraphics[width=\linewidth]{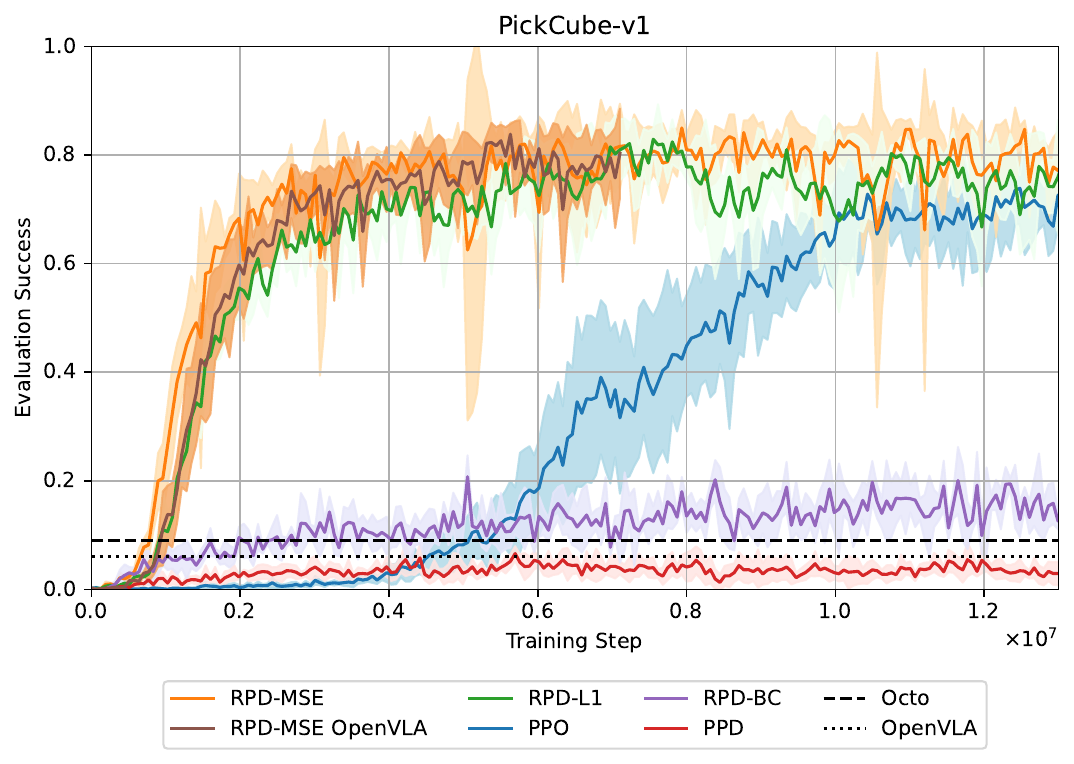}
  \caption{
    Average success rates during training for vanilla PPO, PPD, and three different RPD variants: RPD-MSE, RPD-L1, and RPD-BC, which all distill from Octo over five runs with different seeds.
    Shaded areas indicate standard deviations.
    We also evaluated RPD-MSE on OpenVLA. Due to computing constraints, we could only perform a single training run.
    The performance of the fine-tuned VLAs is indicated by dashed lines.
  }
  \label{fig:algos}
\end{figure}

\autoref{fig:algos} depicts the validation success rates during training for all \rpd{} loss variants and baselines mentioned above.
For reference, it also includes the performance of the VLAs after fine-tuning on the \ms{} dataset presented in~\autoref{tab:tasks}.

The evaluation shows that RPD-MSE outperforms all other variants, closely followed by RPD-L1.
The distilled policy quickly surpasses Octo and OpenVLA to become a refined policy.
RPD-MSE with OpenVLA performs slightly worse than the distillation with Octo.
The PPO baseline also converges to similar success rates as RPD-MSE and \mbox{RPD-L1} of around 80\%, but learns substantially slower and shows much larger fluctuations in the training process.
RPD-BC slightly surpasses Octo's performance threshold but stays at that level and does not refine the policy.
Lastly, PPD fails to lift the policy above the Octo baseline.

The reason why our PPD baseline does not manage to learn the task successfully is likely that it requires the teacher and student policy distributions to be of similar shape for the KL divergence to converge.
However, we found that the action distributions of the VLA policies are sometimes bimodal, which results in non-optimal parameters for the approximated normal distributions that we use to compute the KL divergence in PPD.
We also speculate that in our setup PPD  depends strongly on the teacher's performance.
Lastly, PPD may also be sensitive to PPO hyperparameters, which we have not investigated.
We did, however, modify the parameter $\lambda$, but observed no substantial change in the results.

Even though RPD-BC uses a very similar loss compared to RPD-MSE, it includes a variance component that forces the standard deviation to smaller and smaller values.
This leads to overfitting of the teacher's policy and, thus, does not converge to a considerably better performance.
RPD-BC can therefore be seen as a policy distillation method without refinement.
It may be useful for situations where a close match with the teacher policy is required.

RPD-MSE likely performs better on Octo than on \mbox{OpenVLA} because of the performance difference already present in the foundation models themselves.
Both \mbox{RPD-MSE} and RPD-L1 show quicker and less noisy convergence than the PPO baseline with the same hyperparameters.
This means that RPD efficiently guides the rather inefficient exploration process of the RL agent, leading to larger rewards in less time, which improves sample efficiency and robustness.

\subsection{RPD on Different Tasks}
\begin{figure*}[ht]
  \centering
  \vspace{5.2pt}
  \includegraphics[width=\textwidth]{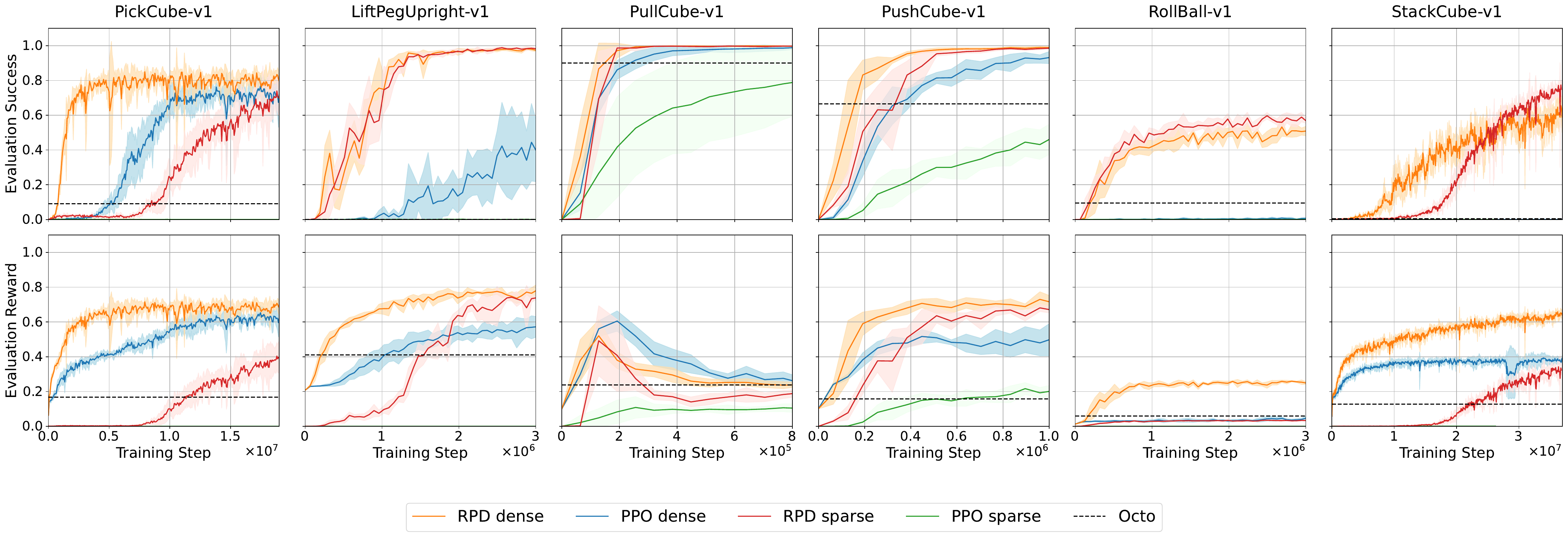}
  \caption{
    Success and reward curves for \rpd{} with Octo and for vanilla PPO for both dense and sparse rewards on the six different \ms{} tasks that are part of the fine-tuning dataset.
    The values are averaged over five runs with different seeds and recorded in an evaluation environment.
    Standard deviations are indicated by shaded areas.
    All training runs were performed with the same hyperparameters from the \ms{} PPO baseline.
    For all tasks, \rpd{} outperforms Octo quickly and converges faster than vanilla PPO.
    In some cases, it even finds good policies when vanilla PPO fails.
    This effect increases for sparse rewards.
  }
  \label{fig:tasks_in}
\end{figure*}

\autoref{fig:tasks_in} shows the training curves for RPD-MSE with Octo on six different manipulation tasks that were part of the training dataset.
In all six cases, RPD learns faster than the PPO baseline and often converges to higher success rates.
This increased performance can also be seen in the reward plots.
For RollBall and StackCube, the PPO baseline fails to solve the tasks with the hyperparameters from \ms{}, whereas RPD achieves at least partial success with the same parameters.
The decreased standard deviation also shows that RPD is generally more robust than the vanilla PPO baseline.
These results demonstrate that RPD improves PPO's training speed consistently and over many tasks while surpassing its final performance. It also refines the distilled policy on dense rewards.

\begin{figure}[t]
  \centering
  \vspace{5.2pt}
  \includegraphics[width=\linewidth]{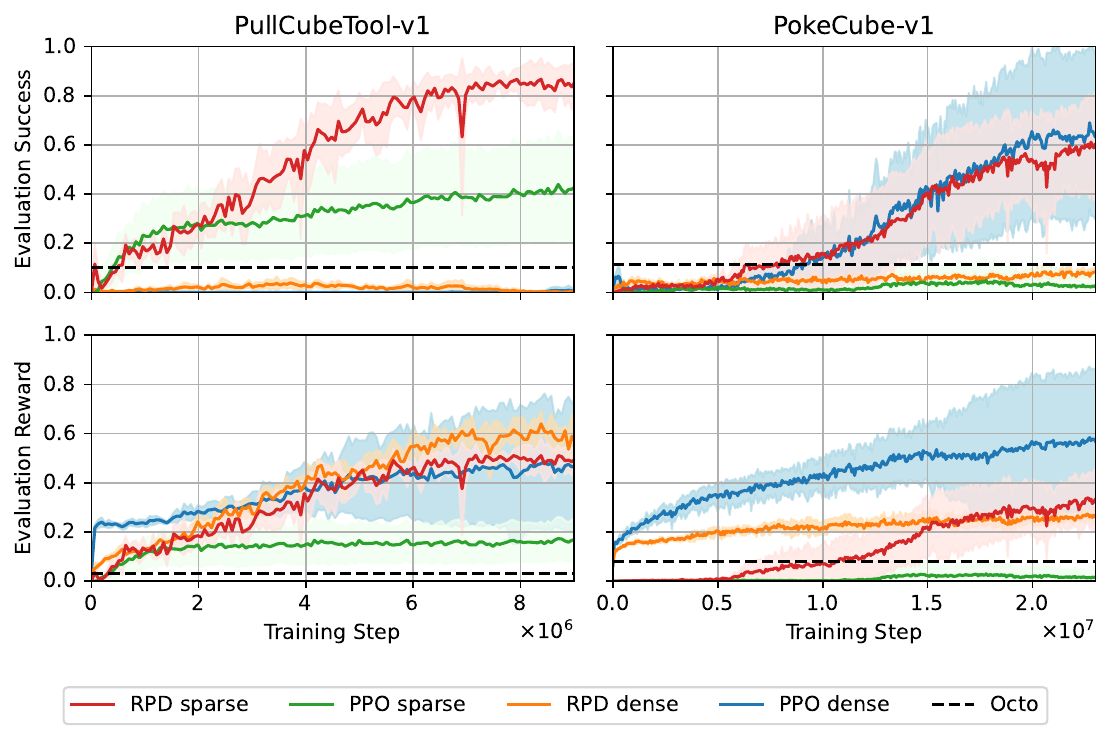}
  \caption{
    Success and reward curves for \rpd{} with Octo and for vanilla PPO on tasks that are \emph{not} part of Octo's fine-tuning dataset.
    The curves show average results for five training runs on different seeds and are recorded in evaluation environments.
    The shaded areas indicate standard deviations.
    Note that the Octo baseline does not represent the correct reward in the dense reward setting as it solves the tasks without tool use. See \autoref{sec:meth} for details.
  }
  \label{fig:tasks_not_in}
\end{figure}

\subsection{Sparse Rewards and Sample Efficiency}
\ms{} defines its sparse rewards as follows:
$-1$ for failed cases, $1$ for success, and $0$ otherwise.
This means that non-zero reward is only given at the end of the episode, which is different from dense rewards where every action usually receives a non-zero reward that is shaped to guide the agent towards a certain behavior.
Sparse reward problems are often a big challenge for RL algorithms, as relevant information about the agent's performance is only obtained once the task has been solved successfully at least one time.
Thus, in the beginning, the agent needs to complete an episode successfully just through random exploration, which makes the training process very inefficient compared to RL with dense rewards.
As a result, it can occur that an agent learns a task perfectly fine with dense rewards but fails in the corresponding sparse reward setting, as can be seen for PPO in \autoref{fig:tasks_in} for some of the tasks.

\autoref{fig:tasks_in} also shows training runs with the six selected \ms{} tasks for sparse rewards.
Note that the resulting reward values cannot be directly compared to the dense reward setting.
The only modification we applied to the hyperparameters from the \ms{} baselines was to set~$\gamma$ from $0.8$ to $0.99$, which avoids vanishing returns for early episode steps.
In all six cases, RPD on sparse rewards consistently outperforms vanilla PPO.
In most tasks, the baseline completely fails to learn the task, whereas RPD succeeds.
It makes sense that RPD can realize its full potential on sparse rewards as it is even more important in this setting to have a teacher policy that enables the student to successfully complete episodes and, thus, retrieve environment rewards that drive the learning process.
In summary, RPD drastically improves sample efficiency for on-policy RL in both sparse and dense reward cases.

\subsection{Generalization to New Environments}
We also evaluated RPD with Octo on tasks that were held out of the VLA fine-tuning dataset to test its generalization capabilities.
Ultimately, RPD is limited by the performance of the teacher and, in this case, by the cross-task generalization of the VLA.
Still, it is worth investigating to what extent the degraded VLA performance will influence the RPD training.
We study this question by looking at task variations, which means they are out-of-distribution with respect to the fine-tuning dataset.
More specifically, we evaluate RPD's performance on the hold-out tasks PullCubeTool and PokeCube, as well as on a variation of the PushCube task with a different camera perspective.

\paragraph{Hold-Out Tasks}
As mentioned in \autoref{sec:meth}, generalization to unseen tasks can be measured based on the performance on the hold-out tasks (see \autoref{tab:tasks}). However, it requires a reinterpretation:
PullCubeTool and PokeCube have to be interpreted as PullCube and PushCube with distracting objects.
For the original versions that require tool usage, we could not observe high generalization capabilities.

The results of our experiments are summarized in \autoref{fig:tasks_not_in}. On the one hand, RPD fails to learn the hold-out tasks on dense rewards as they force the RL agent to use the tool and the VLA policy consequently cannot provide any meaningful actions.
On the other hand, in the sparse reward setting, the RL agent learns the reinterpreted tasks, for which the VLA policy generalizes well.
Therefore, it makes sense that RPD learns these tasks successfully as shown in \autoref{fig:tasks_not_in}.
It outperforms PPO on both tasks even though the reduced task difficulty is the same for both algorithms.

\begin{figure}[t]
  \centering
  \vspace{5.2pt}
  \includegraphics[width=0.48\linewidth]{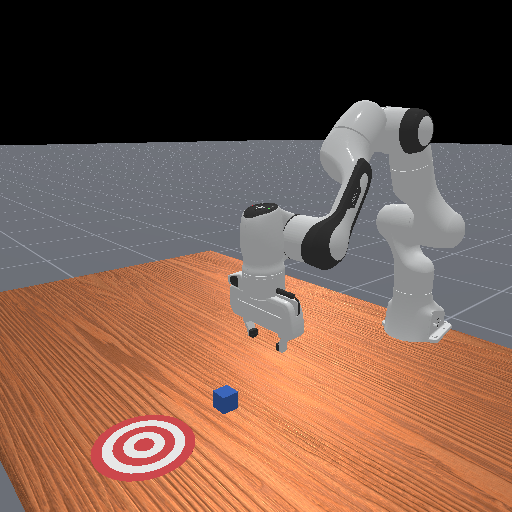}
    \hfill
  \includegraphics[width=0.48\linewidth]{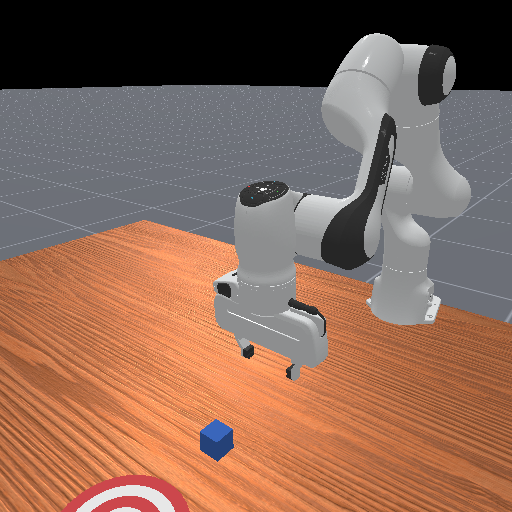}
  \caption{
    The PushCube task with a variation of the camera perspective. The left side shows the perspective used for fine-tuning Octo.
    The right side depicts the perspective we used to test RPD's performance in situations where the VLA needs to generalize from minor setup changes, a task where state-of-the-art VLAs fail.
  }
  \label{fig:camera_perspective}
\end{figure}

\paragraph{Change in Camera Perspective}
VLAs have traditionally been very sensitive to camera adjustments, leading to severe performance degradation~\cite{vla_gen_gap}.
To counteract performance drops, VLAs often require fine-tuning with camera data from the new angle, camera model, etc.
This problem also occurs with our fine-tuned versions of Octo and OpenVLA:
When we only slightly change the camera angle in the PushCube task, as shown in \autoref{fig:camera_perspective}, the success rates of OpenVLA drops from 27\% to 4.5\%. Octo even drops from 67\% to 0\%.
The latter is likely a result of the less powerful vision encoder that is utilized in Octo that may overfit the training data.

\begin{figure}[t]
  \centering
  \vspace{5.2pt}
  \includegraphics[width=\linewidth]{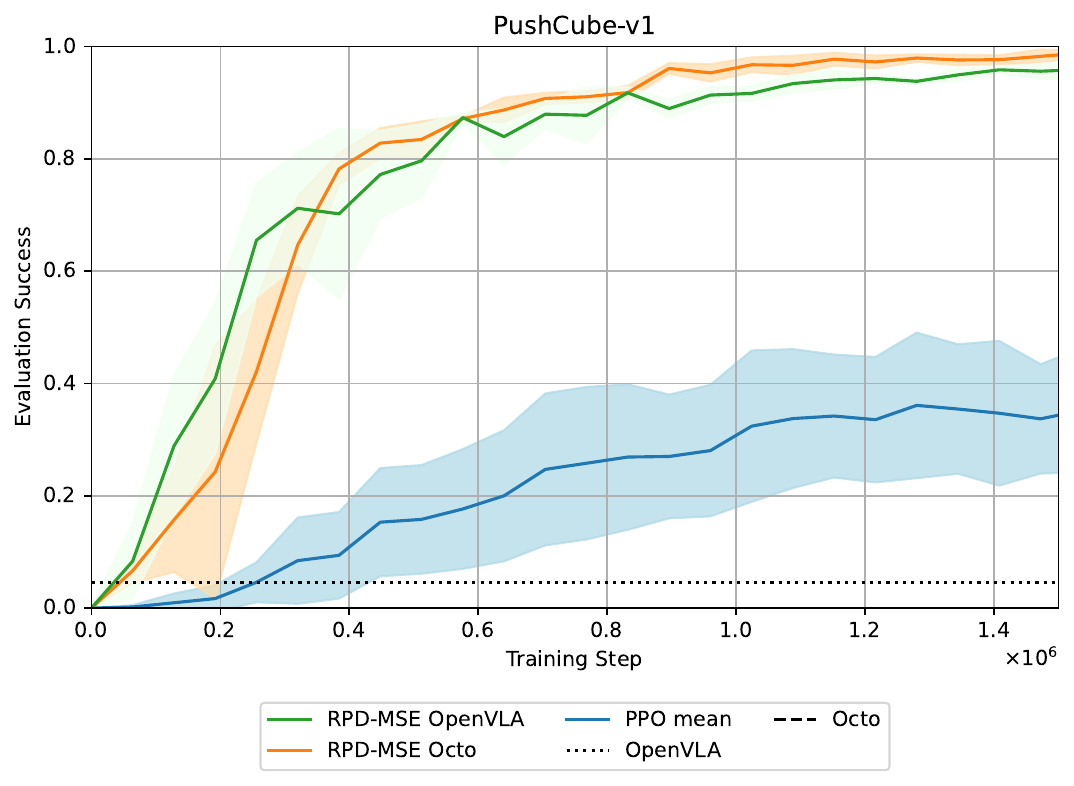}
  \caption{
    Average evaluation success rates of vanilla PPO and of RPD-MSE with both Octo and OpenVLA on a variation of the PushCube task with sparse rewards over five different seeds.
    Shaded areas indicate standard deviations.
    The camera perspective was changed as shown in \autoref{fig:camera_perspective} to investigate the distillation performance of the degraded VLAs.
  }
  \label{fig:angle}
\end{figure}

\autoref{fig:angle} shows the success rates of vanilla PPO and \mbox{RPD-MSE} for both  Octo and OpenVLA on the changed camera perspective from \autoref{fig:camera_perspective} on the PushCube task.
PPO performs similarly on our task variation as in the original PushCube task with sparse rewards, see \autoref{fig:tasks_in}.
This is expected because the agent's visual encoder is learned from scratch in the RL training. Accordingly, there is no bias towards a specific camera angle as long as the provided information is sufficient to solve the task.
Both RPD variants manage to outperform PPO.

Similar to the previous paragraph, where distracting objects are added to the scene, RPD is robust against visual changes despite a considerable drop of the VLA's performance.
The reason why RPD can benefit from degraded VLA actions is because they still guide the exploration in a good direction.
Importantly, low success rates do not necessarily mean that the agent's actions are bad.
For example, the VLA agent may almost complete the task but then perform a wrong action in the end.
In this case, the success rate is zero although the actions are still valuable for a student policy, as can be seen in the evaluation results for RPD.

\subsection{Limitations}
The main limitation of RDP is the performance of the VLA on the task at hand. 
If the VLA cannot provide any meaningful actions due to domain shift or because the task was not included in the training dataset, RPD can be seen as a new exploration component in the loop.
Depending on the resulting action distribution, the VLA may provide bias that is useful for learning the task or could improve the exploration of the environment.
However, if the actions generated by the VLA cannot provide such helpful bias, RPD may also decrease the RL agent's performance. As reported for the training with dense rewards on the hold-out tasks, if the task and no variation of it have ever been seen by the VLA, its actions hinder the PPO training in RPD, and vanilla PPO performs better. Similarly, when PPO is already tuned to a task or the task is very easy, PPO is already sample-efficient, and RPD may only provide limited improvement.

The method is platform-agnostic and can also be employed on a physical setup.
However, despite \rpd{}'s gains in sample efficiency, online RL still needs millions of steps to converge for more complex tasks, which makes it challenging on physical platforms.
For example, sim-to-real methods could be employed to bridge this gap, as described below.

\section{Conclusion and Future Work}
This work introduced \rpd{}, a novel on-policy RL-based approach for distilling task knowledge from generalist VLAs into refined expert policies.
To the best of our knowledge, RPD is the first approach in this direction.
Our experiments show that \rpd{} not only makes the RL training more efficient by guiding its exploration through the actions of the VLA, but also that it can leverage task knowledge from the VLA even in situations where the generalist policy fails, such as when the camera perspective changes.
Thus, \rpd{} can help to improve the usability and accessibility of VLAs.
As new VLAs become available~\cite{pizero}, we expect the performance of \rpd{} to improve even further.

Due to the high data requirements of RL, we evaluated \rpd{} in simulation, which is why we fine-tuned two state-of-the-art generalist VLA policies, Octo and OpenVLA, on simulation datasets.
As a side contribution, our work shows that VLAs can be successfully utilized in experiments in simulated environments.
This provides a promising perspective for scientists who have no access to a physical robot setup to conduct research on VLAs.
An interesting direction for future research is to combine RPD with sim-to-real methods~\cite{simtoreal} for deployment on real robots.

Another promising line of research is to use the distilled expert policies for collecting new training data. Following the method proposed by Xu \emph{et al.}~\cite{rldg} to fine-tune VLAs, it will be possible to improve generalist policies without collecting any additional human demonstration data.

\section*{Acknowledgement}
This work is supported by the project GeniusRobot funded by the German Federal Ministry of Education and Research (BMBF grant no.~01IS24083).
The authors acknowledge the HPC resources provided by the Erlangen National HPC Center (NHR@FAU) under the BayernKI project no.~v108be.

\bibliographystyle{IEEEtran}
\bibliography{bibliography.bib}

\end{document}